\documentclass[runningheads]{llncs}
\usepackage[numbers,sort&compress,sectionbib]{natbib}
\usepackage{graphicx}
\usepackage{array}

\usepackage{amssymb}
\usepackage{amsmath,nccmath}
\usepackage{float}
\usepackage{subfig}
\usepackage{mathtools}
\usepackage{multirow}

\begin{document}
\title{A generalised form for a homogeneous population of structures using an overlapping mixture of Gaussian processes}
	
\titlerunning{A generalised form for a homogeneous population of structures}
\author{Tina A. Dardeno\inst{1} \and
Lawrence A. Bull\inst{2}\and
Nikolaos Dervilis\inst{1} \and
Keith Worden\inst{1}}
\authorrunning{T.A. Dardeno \emph{et al.}}
\institute{Dynamics Research Group, Department of Mechanical Engineering, University of Sheffield, Mappin Street, Sheffield S1 3JD, UK \and
The Alan Turing Institute, The British Library, London NW1 2DB, UK\\
\email{t.a.dardeno@sheffield.ac.uk}}

\maketitle     

\begin{abstract}
	 Reductions in natural frequency are often used as a damage indicator for structural health monitoring (SHM) purposes. However, fluctuations in operational and environmental conditions, changes in boundary conditions, and slight differences among nominally-identical structures can also affect stiffness, producing frequency changes that mimic or mask damage. This variability has limited the practical implementation and generalisation of SHM technologies. The aim of this work is to investigate the effects of normal variation, and to identify methods that account for the resulting uncertainty. \\
	 
	 This work considers vibration data collected from a set of four healthy full-scale composite helicopter blades. The blades were nominally-identical but distinct, and slight differences in material properties and geometry among the blades caused significant variability in the frequency response functions, which presented as four separate trajectories across the input space. In this paper, an overlapping mixture of Gaussian processes (OMGP), was used to generate labels and quantify the uncertainty of normal-condition frequency response data from the helicopter blades. Using a population-based approach, the OMGP model provided a generic representation, called a \emph{form}, to characterise the normal condition of the blades. Additional simulated data were then compared against the form and evaluated for damage using a marginal-likelihood novelty index.

\end{abstract}

\section{Introduction}
	A number of challenges affect the practical implementation and generalisation of structural health monitoring (SHM) technologies. Features that are sensitive to damage may also be sensitive to operational and environmental fluctuations or changes in boundary conditions, which may cause issues with discerning between damaged states and normal variations. In addition, difficulties arise when generalising between different but nominally-identical (i.e., \emph{homogeneous} \cite{Bull_1,Bull_2}) structures, such as helicopter blades, as small variations in the internal structure and material properties can present as changes in dynamics. These changes, like those caused by environmental and other fluctuations, may be erroneously flagged as damage, or mask subtle damage in a structure. Collecting comprehensive data covering a range of operational and damaged conditions from each individual structure is typically not feasible. In such cases, population-based SHM (PBSHM) can be used to make inferences between different members of a \emph{population}, i.e., groups of similar systems \cite{Bull_1,Bull_2,gosliga2021foundations,gardner2021foundations,Bull_3}. 
	
	PBSHM seeks to transfer valuable information across similar structures. In certain cases, the behaviour of the group can be represented using a general model. Bull \emph{et al.}\ used Gaussian process (GP) regression of real \cite{Bull_1,Bull_2}, and imaginary \cite{Bull_2}, frequency response functions (FRFs) to develop a generic representation, called a \emph{form}, of a population of nominally-identical, but slightly different, eight degree-of-freedom (DOF) systems. The population was comprised of multiple simulated healthy and damaged systems \cite{Bull_1,Bull_2}, and one experimental rig, which was used to test the form in \cite{Bull_1}. Normal-condition data from several members of the population were used to train the form, and the posterior predictive distributions of the GP were used to aid novelty detection across the population, using the Mahalanobis squared-distance novelty index \cite{WORDEN2000647}. The form was found to predict accurately the healthy and damaged states across the population \cite{Bull_1,Bull_2}. In addition, an overlapping mixture of Gaussian processes (OMGP) \cite{LAZAROGREDILLA20121386} was used in \cite{Bull_1,Bull_3} to infer multivalued wind-turbine power-curve data, with categorisation of the data performed in an unsupervised manner. Novelty detection was performed using the negative log of the marginal likelihood of the OMGP, and correctly identified \emph{good} and \emph{bad} power curves for the majority of data tested \cite{Bull_1,Bull_3}.

	The current work is focussed on applying structural health monitoring (SHM) to a homogeneous population of structures for the purpose of damage detection. (As opposed to a \emph{heterogeneous} population \cite{gosliga2021foundations,gardner2021foundations}, where the members are more disparate, such as different designs of a suspension bridge). Data were collected from four healthy, nominally-identical helicopter blades. FRFs were computed from the measured time-domain data, and these data were used to develop a form for the blades using an (unsupervised) OMGP. The posterior predictive distributions from the OMGP were later applied to a novelty detection approach to evaluate new (simulated) FRF data against the form. As such, this work builds upon that performed in \cite{Bull_1,Bull_2}, by extending the concept of a population form to a set of nominally-identical helicopter blades with a range of undamaged-condition FRF data, using mixtures of probabilistic regression models \cite{Bull_1,Bull_3,LAZAROGREDILLA20121386}, to account for the large variability in the FRFs of the blades. The significant discrepancies in the experimental data set are representative of the differences in the blades in practice, with variability arising from small variations in the internal structure and material properties of the helicopter blades, as well as slight changes in boundary conditions. The small population of helicopter blades provided a useful data set to develop and test the techniques presented herein, such that these methods can be later applied to other homogeneous populations such as a wind farm.	
	
	The layout of this paper is as follows. Section 2 provides a brief overview of the OMGP theory and Section 3 introduces the data set used in the analyses. Section 4 discusses how a generalised normal condition was determined for the helicopter blades using an (unsupervised) OMGP. Finally, Section 5 demonstrates how the OMGP can be used to inform damage detection, by computing the marginal likelihood for a series of (simulated) damaged FRFs.
	
\section{Overlapping mixtures of Gaussian processes (OMGP)}

	In this work, a generalised normal condition (called a \emph{form}), was determined for a small population of nominally-identical helicopter blades using an (unsupervised) overlapping mixture of Gaussian processes (OMGP). A brief overview of the OMGP technique is provided below. A more thorough discussion of the theory can be found in \cite{LAZAROGREDILLA20121386,Bull_1,Bull_3}.
	
	The OMGP is a useful and flexible method for developing the population form, as no assumptions are made regarding data classification (although the number of trajectories across the input-space must be known \emph{a priori}). The OMGP \cite{LAZAROGREDILLA20121386,tay2008modelling,Bull_1,Bull_3}, describes a data set by evaluating each observation using one of \emph{K} latent functions plus additive noise,
	
		\begin{equation}
			y_i^{(k)} = \{f^{(k)}(x_i) + \epsilon_i\}_{k=1}^{K} 
			\label{eq:latentfuncOMGP}
		\end{equation}

	The OMGP is unsupervised, i.e., the labels that assign the data to a given function are unknown. Therefore, another latent variable is introduced, $ \mathbf{Z} $, which is a binary indicator matrix. The matrix $ \mathbf{Z} $ is entirely populated with zeroes, except for one non-zero entry per row. The non-zero entries, $ \mathbf{Z}[i,k] \neq 0 $, indicate that observation $ i $ was generated by function $ k $. The likelihood of the OMGP is therefore written as \cite{LAZAROGREDILLA20121386,Bull_3}, 
	
		\begin{equation}
			p(\mathbf{y} \;|\; \{\mathbf{f}^{(k)}\}_{k=1}^{K}, \mathbf{Z}, 	\mathbf{x}) =
			\prod_{i,k = 1}^{N,K}
			p{(y_i \;|\; {f}^{(k)}(x_i))}^{\mathbf{Z}[i,k]}
			\label{eq:likelihoodOMGP}
		\end{equation}
	
	Prior distributions are then placed over the latent functions and variables, such that \cite{Bull_3},
	
		\begin{equation}
			P(\mathbf{Z}) = \prod_{i,k = 1}^{N,K}  	{\boldsymbol{\varPi}[i,k]}^{\mathbf{Z}[i,k]} 
			\label{eq:priorOMGP1}
		\end{equation}
	
		\begin{equation}
			f^{(k)}(x_i) \sim \mathcal{GP}(m^{(k)}(x_i),k^{(k)}(x_i,x_j)), \epsilon_i \sim \mathcal{N}(0,\sigma^2)  
			\label{eq:priorOMGP2}
		\end{equation}
	
	\noindent where $ P(\mathbf{Z}) $ is the prior over the indicator matrix $ \mathbf{Z} $, and $ \boldsymbol{\varPi}[i,:] $ is a histogram over the $ K $ components for the $ i $th observation, and $ \sum_{k=1}^{K}\boldsymbol{\varPi}[i,k] = 1 $ \cite{Bull_3}. The terms in Eq.\ (\ref{eq:priorOMGP2}) are independent GP priors over each latent function $ f^{(k)} $ with distinct mean $ m^{(k)}(x_i) $ and kernel $ k^{(k)}(x_i,x_j) $ functions \cite{Bull_3}. A shared hyperparameter $ \sigma $, is used to define the prior over the noise variances, to reduce the number of latent variables in the calculations \cite{Bull_3}. This work used a modal damping-based FRF estimation as the mean function for the GP regression. 
	
	\subsection{Mean function}
	
		To incorporate prior knowledge of the dynamic properties of the blades, a non-zero mean function was introduced, via an estimation of the accelerance FRF, 
		
			\begin{equation}
				\mathbf{H}_{ij}(\omega) = -\omega^2 \sum_{k=1}^{n} 	\frac{\text{A}_{ij}^{(k)}}{\omega_{nk}^2-\omega^2+2i\zeta_k\omega\omega_{nk}}
				\label{eq:modalFRF}
			\end{equation}
	
		\noindent where $ \text{A}_{ij}^{(k)} $ is the residue for mode $ k $, defined as the product of the mass-normalised mode shapes at locations $ i $ and $ j $ ($ \text{A}_{ij}^{(k)} = \psi_{ik}\psi_{jk} $) \cite{wordennonlinearity}. The natural frequency associated with mode $ k $ is $ \omega_{nk} $, and the modal damping associated with mode $ k $ is $ \zeta_k $ \cite{wordennonlinearity}. Modal damping, residue, and natural frequency provided the mean-function hyperparameters. 
	
	\subsection{Variational inference and Expectation Maximisation (EM)}
	
		Exact computation of the posterior distribution $ p(\{\mathbf{f}^{(k)}\}_{k=1}^{K},\mathbf{Z} \;|\; \mathbf{x},\mathbf{y}) $ is intractable; therefore, a variational inference \cite{VIreview} and Expectation Maximisation (EM) scheme was employed to estimate the posterior. The scheme consisted of alternating between the E-step (mean-field updates to the variational distribution, while keeping hyperparameters fixed), and M-step (optimisation of hyperparameters while keeping the distribution fixed), until the convergence of a lower bound, which coincided with maximising the marginal likelihood of the model. A detailed review of the theory can be found in  \cite{LAZAROGREDILLA20121386,Bull_1,Bull_3}.
	
		In this work, constrained nonlinear optimisation was performed via the \emph{fmincon} function in MATLAB, and appropriate bounds were applied to the hyperparameters given prior knowledge of the modal characteristics of the blades.

	\subsection{Predictive equations}
	
		Once learnt, the OMGP form can be used to estimate the latent variables and functions to assist damage detection. For training data $ \mathcal{D} $, the \emph{maximum a posteriori} (MAP) estimate can be used to categorise observations according to the most likely component $ k $, via \cite{Bull_3}, 
	
			\begin{equation}
				\hat{k}_i = \underset{k}{\operatorname{argmax}} 	\{\hat{\boldsymbol{\Pi}}[i,k]\}
				\label{eq:k1} 
			\end{equation}
		
			\noindent The MAP class component for new data $ \hat{k}_* $ is then defined as \cite{Bull_3}, 
		
			\begin{equation}
				\hat{k}_* = \underset{k_*}{\operatorname{argmax}} \{p({k}_* 	\;|\; \bold{x}_*,\bold{y}_*,\mathcal{D})\}
				\label{eq:k2} 
			\end{equation}

		\noindent where, for a set of test data, the posterior predictive class component $ k_* $ is expressed using Bayes Rule (note that to classify new data according to $ k_* $, both $ \mathbf{x_*} $ and $ \mathbf{y_*} $ must be observed \cite{Bull_3}),
		
			\begin{equation}
				p({k}_* \;|\; \bold{x}_*,\bold{y}_*,\mathcal{D}) =
				\frac{{p(\bold{y}_* \;|\; \bold{x}_*,k,\mathcal{D})p(k_*)}}
				{{p(\bold{y}_* \;|\; \bold{x}_*,\mathcal{D})}}
				\label{eq:PredictiveEqs}
			\end{equation}

		\noindent The numerator of Eq.\ (\ref{eq:PredictiveEqs}), or unnormalised posterior, is defined as \cite{Bull_3},
	
			\begin{equation}
				p(\bold{y}_* \;|\; \bold{x}_*,k,\mathcal{D})p(k_*)
				\overset{\Delta}{=} 
				\mathcal{N}(\bold{y}_* \;|\;
				\boldsymbol{\mu}_*^{(k)},\boldsymbol{\varSigma}_*^{(k)})
				\boldsymbol{\varPi}[*,k]
				\label{eq:PredictiveEqs1}
			\end{equation}
	
		\noindent and the denominator of Eq.\ (\ref{eq:PredictiveEqs}), which is the marginal likelihood, is \cite{Bull_3},
	
			\begin{equation}
				\begin{split}
					p(\bold{y}_* \;|\; \bold{x}_*, \mathcal{D}) & \approx 	\sum_{k=1}^{K} \boldsymbol{\varPi}[*,k] \int p\left(\bold{y}_* \;|\; \mathbf{f}^{(k)},\bold{x}_*, \mathcal{D}\right)q\left(\mathbf{f}^{(k)} \;|\;\mathcal{D}\right)\text{d}\mathbf{f}^{(k)} \\
					& = \sum_{k=1}^{K}\mathcal{N}\left(\bold{y}_* \;|\; 	\boldsymbol{\mu}_*^{(k)},\boldsymbol{\varSigma}_*^{(k)}\right)\boldsymbol{\varPi}[*,k] \\
					\label{eq:PredictiveEqs2}
				\end{split}
			\end{equation}
	
		\noindent where \cite{Bull_3},

			\begin{equation}
				\begin{split}
					& \boldsymbol{\mu}_*^{(k)} \overset{\Delta}{=} 	\mathbf{m}_*^{(k)} + \mathbf{K}^{(k)}_\mathbf{{x_*x}}\left(\mathbf{K}^{(k)}_\mathbf{{xx}} + \mathbf{B}^{(k)-1}\right)^{-1}\left(\mathbf{y} - \mathbf{m}^{(k)}\right) \\
					& \boldsymbol{\Sigma}_*^{(k)} \overset{\Delta}{=} 	\mathbf{K}^{(k)}_\mathbf{{x_*x_*}} - \mathbf{K}^{(k)}_\mathbf{{x_*x}}\left(\mathbf{K}^{(k)}_\mathbf{{xx}} + \mathbf{B}^{(k)-1}\right)^{-1} \mathbf{K}^{(k)}_\mathbf{{xx_*}} + \mathbf{R}_*^{(k)} \\
					& \mathbf{R}_*^{(k)} \overset{\Delta}{=} \sigma^2 	\mathbf{I}_\mathbf{M}\\
					\label{eq:PredictiveEqs3}
				\end{split}
			\end{equation}
	
		\noindent The quantity  $ \mathbf{B}^{(k)} $ is an $ N \times N $ diagonal matrix with elements,
	
			\begin{equation}
				\mathbf{B}^{(k)} = 		\text{diag}\left(\left\{\frac{\boldsymbol{\hat{\varPi}}[1,k]}{\sigma^2}\;,\;...\;,\; \frac{\boldsymbol{\hat{\varPi}}[N,k]}{\sigma^2}\right\}\right)
				\label{eq:Estep3}
			\end{equation}
		
		The prior mixing proportion for new observations, $ \boldsymbol{\varPi}[*,k] $, weights each component equally \emph{a priori}, such that the sum of the weights is equal to 1, or $ \boldsymbol{\varPi}[*,k] = 1/K $ \cite{Bull_3}. The predictive equations for the OMGP are quite similar to those for the conventional GP \cite{Bull_3}, with the exception of the noise component for the training data $ \mathbf{B}^{(k)-1} $ which is scaled for each sample according to the inverse of the mixing proportion  $ \boldsymbol{\hat{\varPi}}[i,k]^{-1} $ \cite{LAZAROGREDILLA20121386,Bull_3}. As such, each sample contribution is weighted with respect to its posterior predictive component in the mixture \cite{Bull_3}. 
	
	\subsection{Practical implementation of the described technology}
	
		In this work, a population form was developed for the helicopter blades using an OMGP, and new (simulated) FRF data were used to evaluate the sensitivity of the form for novelty detection. There were four healthy helicopter blades and therefore four classes ($ K = 4 $). Because the new data had an equal chance of belonging to any of the four classes, the prior mixing proportion weighted each class equally, with $ \boldsymbol{\varPi}[*,k] = 0.25 $. The marginal likelihood (evidence) defined in Eq.\ (\ref{eq:PredictiveEqs2}) was used as a novelty index, and provided an averaged likelihood of new data belonging to any of the $ k $ classes. For computational ease and to avoid numerical underflow, the calculations were performed in log space. Specifically, negative log quantities were used, such that the visual presentation of data novelty was consistent with the Mahalanobis squared-distance (MSD) novelty index from \cite{Bull_1,Bull_2,WORDEN2000647}. As such, data below a certain threshold were considered normal or \emph{inyling}, and data above the threshold were considered novel or \emph{outlying}.
	
		A suitable novelty-detection threshold was determined using a similar method as with the MSD in \cite{Bull_1,Bull_2,WORDEN2000647}. Bootstrap-sampling was used, where 1000 samples were randomly selected from the normal-condition data used to train the form. The marginal likelihood was then calculated for the samples, for a large number of trials. The critical value was the threshold with a certain percentage of the calculated values below it. This work used a 99\% confidence interval, or $ 2.58\sigma $.

\section{Data set summary}
	
\begin{figure}[h]
			\vspace{-0.5cm}
			\centering
			\subfloat[\label{fig:FRFs_allblades_fixed}]{\includegraphics[width=0.5\textwidth]{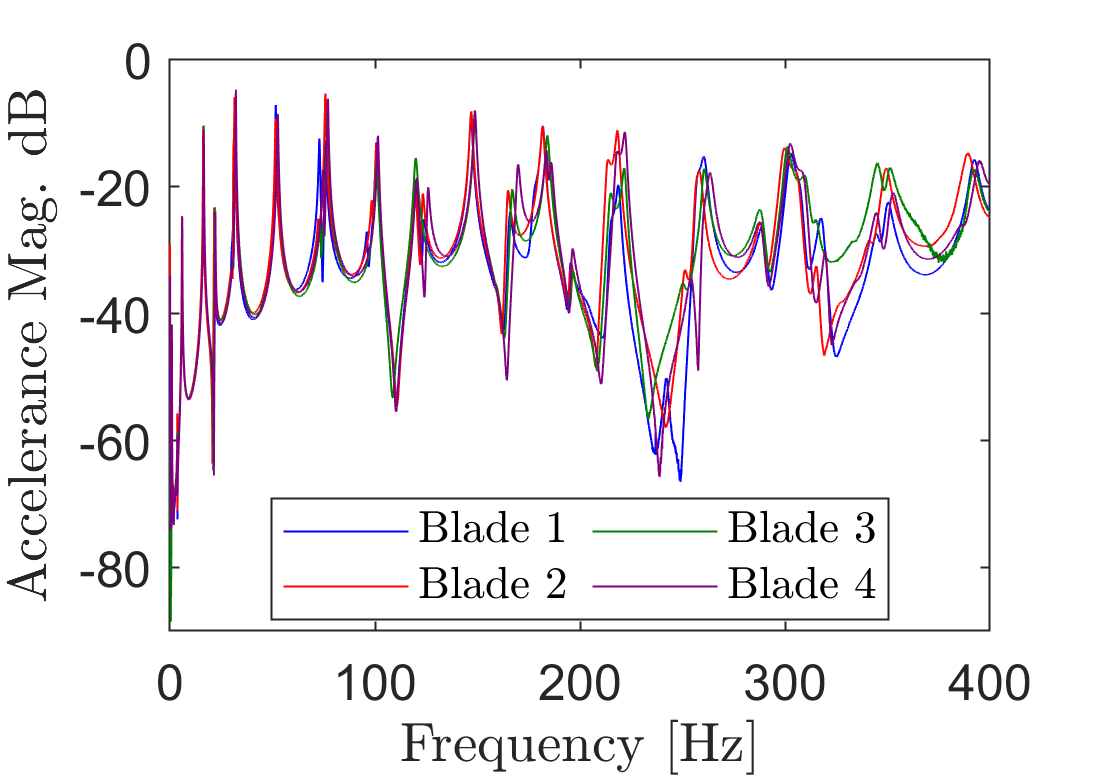}}
			\subfloat[\label{fig:FRFs_allblades_fixed_zoomed}]{\includegraphics[width=0.5\textwidth]{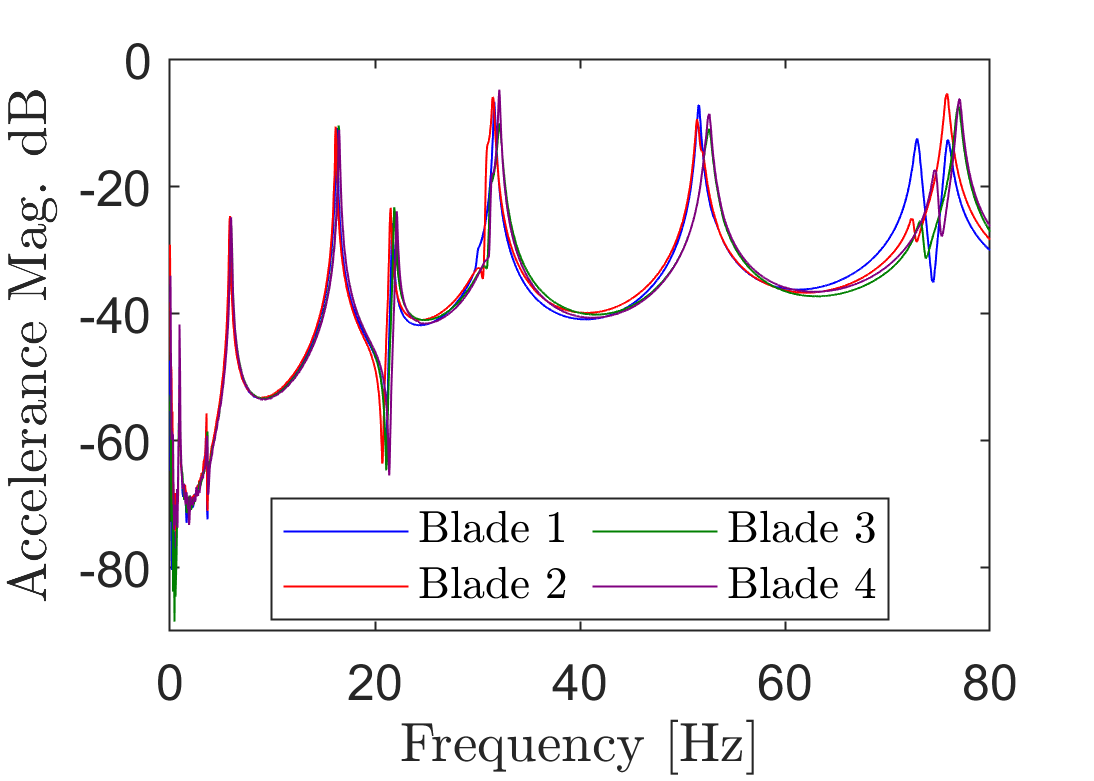}} \\
			\caption{FRF magnitudes for the blades in fixed-free (a) full bandwidth and (d) first 80 Hz.}
			\vspace{-0.5cm}
		\end{figure}

	To develop the OMGP form, FRFs were computed from vibration data that were collected at ambient temperature on four healthy, full-scale composite helicopter blades in a fixed-free boundary condition. An electrodynamic shaker attached near the blade root supplied a continuous random excitation designed to excite the blades up to 400 Hz in the flapwise direction. Data were collected via multiple 100 mV/g accelerometers located on the underside of each blade, although this work only considered vibration data obtained at an accelerometer near the blade tip. Twenty linear averages were obtained in the frequency domain to reduce noise effects. The averaged FRFs for each blade in the fixed-free condition are shown in Figures \ref{fig:FRFs_allblades_fixed} and \ref{fig:FRFs_allblades_fixed_zoomed}. Figure \ref{fig:FRFs_allblades_fixed} shows the full measured 400 Hz bandwidth and Figure \ref{fig:FRFs_allblades_fixed_zoomed} shows only modes below 80 Hz. 
	
	Figures \ref{fig:FRFs_allblades_fixed} and \ref{fig:FRFs_allblades_fixed_zoomed} show considerable variability among the blades, with discrepancies as high as 6.3 Hz (via peak-picking). To develop the form, four distinct data trajectories were assumed, consistent with the number of blades contributing to the data set, to account for horizontal variability among the blades. Further details regarding the data set and experimental setup can be found in \cite{DardenoIWSHM1}.

\section{Development of a population form using an overlapping mixture of Gaussian processes (OMGP)}
	
	An OMGP assumes that the number of classes are known \emph{a priori}, but does not assume which data belong to a given class. In this case, it was assumed known that four helicopter blades contributed to the data (i.e., $ K = 4 $). The data were then classified and fitted by approximating the intractable posterior distribution $ p(\{\mathbf{f}^{(k)}\}_{k=1}^{K},\mathbf{Z} \;|\; \mathbf{x},\mathbf{y}) $ via the OMGP approach described in \cite{LAZAROGREDILLA20121386}, using the variational inference and EM scheme laid out in \cite{LAZAROGREDILLA20121386,tay2008modelling,Bull_1,Bull_3}.
		
	To develop the form, a narrow frequency band was selected between 48 and 56 Hz, containing the fifth bending mode of the blades. Closer inspection of the data revealed a likely second mode in the same frequency range; however, the modes were so closely-spaced that they could conceivably be a slightly distorted single mode. An SDOF assumption was imposed, and an SDOF mean function was applied to the GPs. In other words, the analysis was treated as a grey box, in that some physics were imposed via the mean function. Simplifying the analysis to include a single mode in the band of interest was a reasonable assumption, as most of the physics (and variation) were accounted for with the SDOF mean function, and the remaining distortion was resolved via the black box component of the GP. (Note that the decision to focus on a narrow frequency band was made to simplify the analysis while demonstrating the proposed technology, although the form could be developed over a larger band and with an MDOF assumption).
		
	Real and imaginary parts of the FRF were fitted separately via the OMGP approach. (Note that several conference papers related to this research \cite{DardenoIWSHM1,DardenoIWSHM2}, provided a preliminary approach where magnitude was fitted directly using a single GP. Although the resulting form was useful for damage detection, the possible negative samples do not make statistical sense and were therefore not employed in later work.) Prior to fitting the OMGP, the real and imaginary parts of the FRFs were copied 20 times, and random Gaussian noise with a magnitude equal to 5\% of the absolute peak value of each FRF was added to each copy. The data copies were then concatenated into a vector, and 600 training points were randomly selected from the vector. The real part was fitted first, and the FRF model from Eq.\ (\ref{eq:modalFRF}) was used to generate a mean function, with the modal damping, residue, and natural frequency used as mean-function hyperparameters. Upper and lower bounds were applied to constrain the hyperparameter optimisation, based on prior knowledge of the blades and visual inspection of the data (e.g., the natural frequency was bounded between 40 and 60 Hz). In addition, the sign of the data ($ + / - $) was assumed to be known \emph{a priori}, and can easily be determined via data inspection. 
		
\begin{figure}[htb]
			\vspace{-0.5cm}
			\centering
			\subfloat[\label{fig:blades_GP_real_EX2}]{\includegraphics[width=0.5\textwidth]{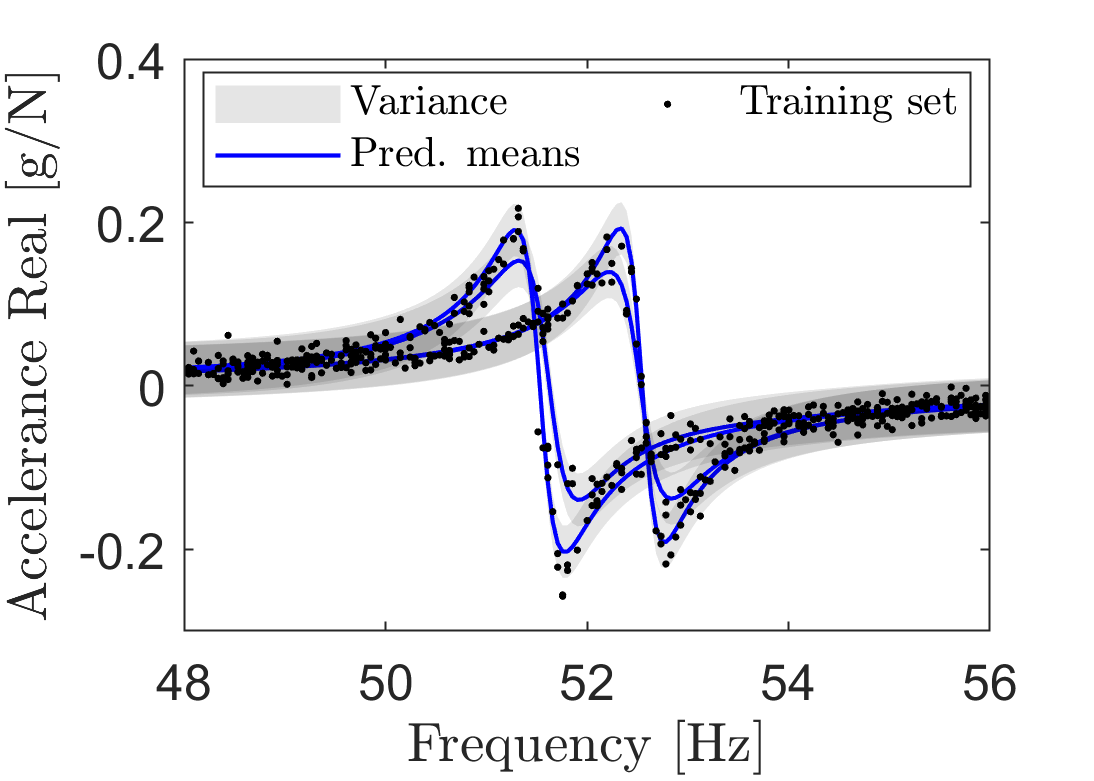}} 
			\subfloat[\label{fig:blades_GP_imag_EX2}]{\includegraphics[width=0.5\textwidth]{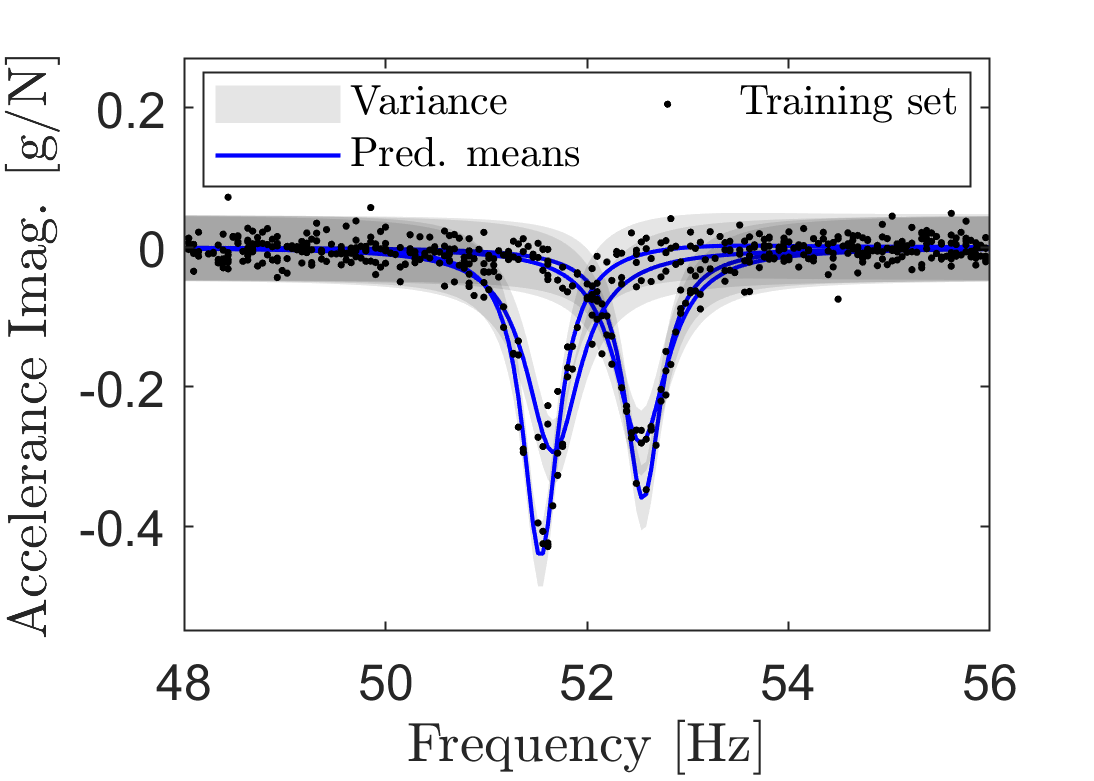}} \\ [-0.5ex]
			\caption{OMGP predictions using (a) real and (b) imaginary FRF data.}
			\vspace{-0.5cm}
		\end{figure}

	Because the real part of the FRF was more separable than the imaginary part (indeed, visual inspection of the FRFs showed that four distinct trajectories in the training data were harder to decipher from the imaginary part versus the real part), the OMGP performed much better for a variety of initialisations on the real data compared to the imaginary data. As such, the real data were fitted first, and an optimal solution was found by randomly initialising the process (i.e., via prior guesses for the hyperparameters), ten times while keeping the training set fixed, and choosing the solution with the maximum lower bound and therefore the highest marginal likelihood. The initialisation parameters were chosen randomly, but were restricted to reasonable intervals in the same manner as the hyperparameters were bounded during the optimisation. Once learnt, the optimised hyperparameters for the real data were used to initialise the OMGP for the imaginary data. The posterior predictive mean of the OMGP is plotted with the mean function, variance, and training data for the real and imaginary FRFs in Figures \ref{fig:blades_GP_real_EX2} and \ref{fig:blades_GP_imag_EX2}. 
		
	Visual inspection of Figures \ref{fig:blades_GP_real_EX2} and \ref{fig:blades_GP_imag_EX2} show that the OMGP provided a reasonable fit for the FRF data, and the majority of the training data were enclosed by the variance bounds. Once fitted, the OMGP form could be used to assist damage detection, by comparing new data against the form, with consideration for normal variability among the helicopter blades.
		
\vspace{12pt} 
\section{Novelty detection via the form}
		
	One practical application of the OMGP form is for damage detection, where unseen data can be compared against the posterior predictive distributions of the OMGP and evaluated for novelty. This section shows how novelty detection could be accomplished, using a technique similar to that involving the MSD in \cite{Bull_1,Bull_2,WORDEN2000647}, but using the evidence from Eq.\ (\ref{eq:PredictiveEqs2}) as a novelty index. Because the real and imaginary parts of the FRFs were fitted separately, the sum of the (negative log) marginal likelihood from each fit provided the novelty index. A Monte Carlo approach (as in \cite{WORDEN2000647}) was used to identify a suitable normal-condition threshold. 
		
	To evaluate the sensitivity of the form to changes in natural frequency, real and imaginary SDOF FRFs were synthesised using Eq.\ (\ref{eq:modalFRF}) to simulate damage-related stiffness reduction, which may present as a downward shift in natural frequency. FRFs were generated for each blade using modal information from the experiments and natural frequency incrementally decreased by 0.5\% for a maximum reduction of 3.5\%. The simulated FRFs were copied 1000 times and random Gaussian noise with a magnitude equal to 5\% of the absolute peak value of the FRF was added to each copy. The negative log marginal likelihood was then computed for each copy, with the novelty index equal to the negative log sum of the evidence for the real and imaginary parts. Likewise, the experimental data were copied 1000 times and random Gaussian noise with a magnitude equal to 5\% of the absolute peak value of the FRF was added to each copy. As with the simulated data, the copied experimental data (with added noise), were tested against the OMGP form, to evaluate the performance of the novelty detector for normal-condition data. 
		
\begin{figure}[h!]
			\centering
			\subfloat[\label{fig:Synth1_2}]{\includegraphics[width=0.5\textwidth]{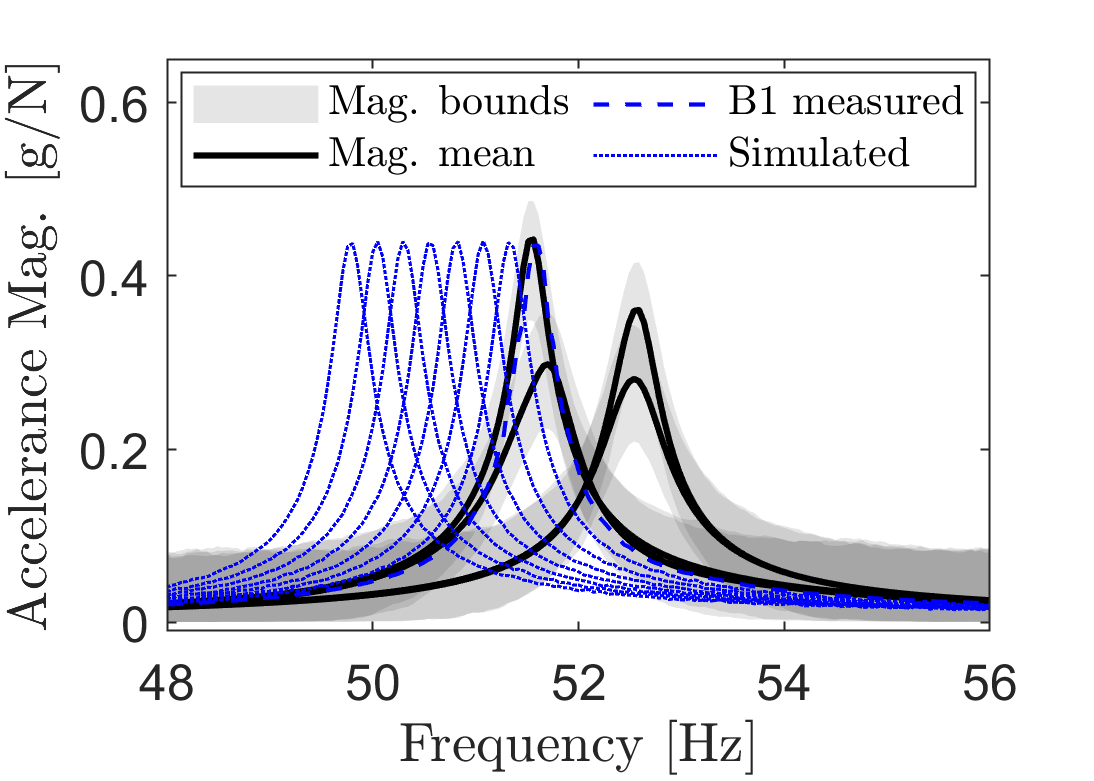}} 
			\subfloat[\label{fig:Synth2_2}]{\includegraphics[width=0.5\textwidth]{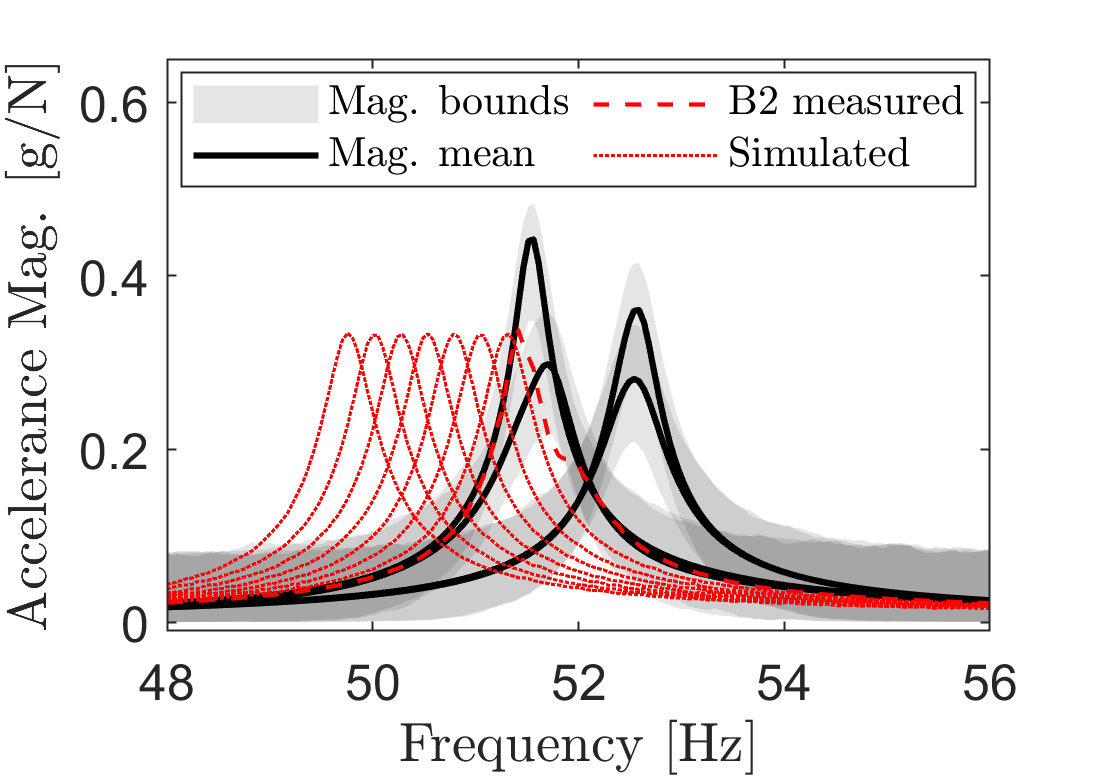}} \\ [-3ex]
			\subfloat[\label{fig:Synth3_2}]{\includegraphics[width=0.5\textwidth]{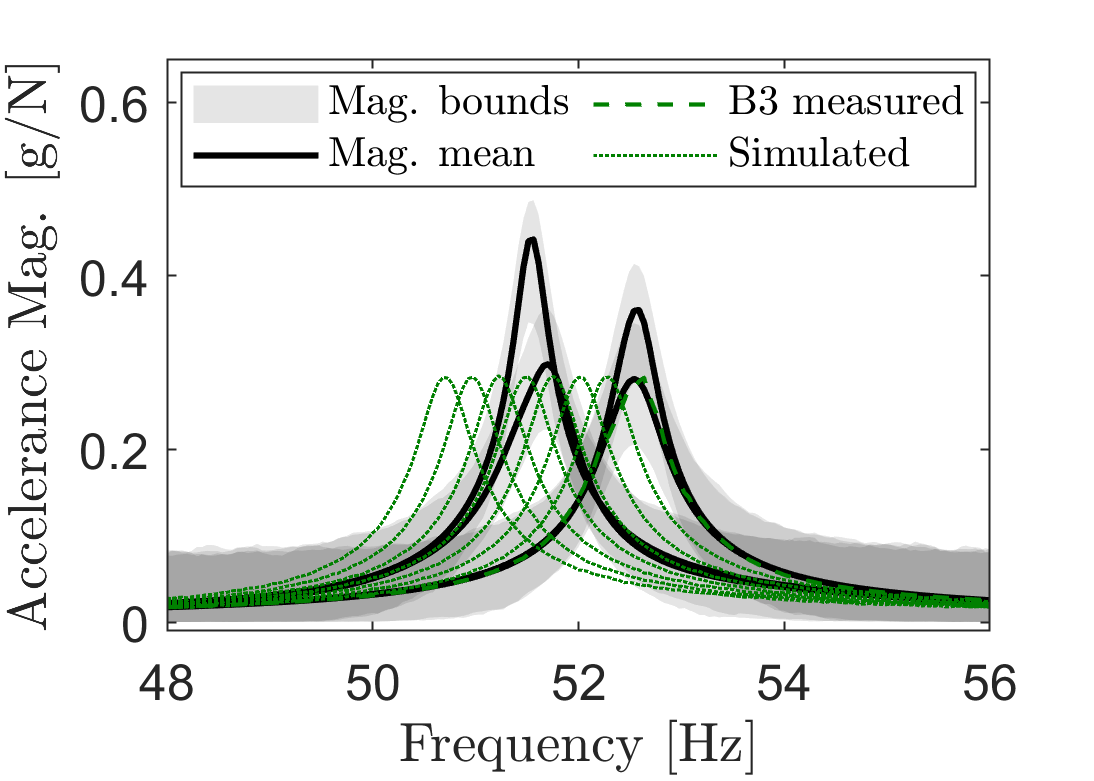}} 
			\subfloat[\label{fig:Synth4_2}]{\includegraphics[width=0.5\textwidth]{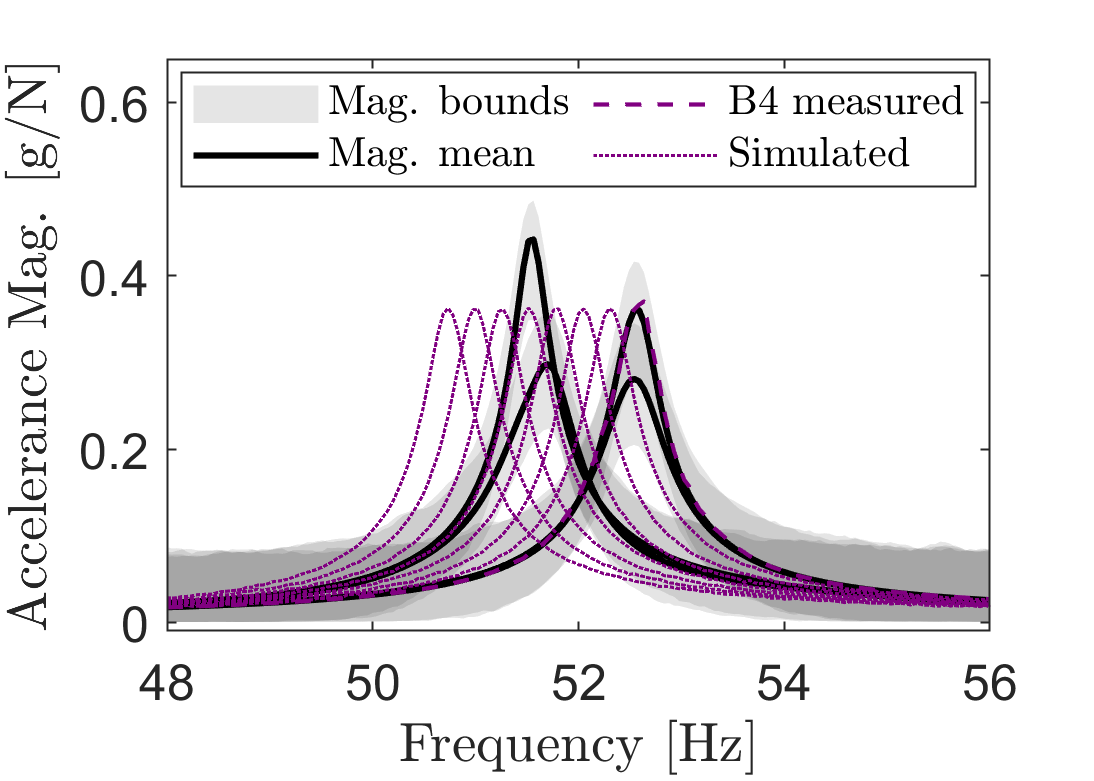}} \\
			\caption{Synthesised FRFs with incrementally decreasing natural frequency for (a) Blade 1, (b) Blade 2, (c) Blade 3, and (d) Blade 4, superimposed on the magnitude of the predicted OMGP.}
			\vspace{-0.5cm}
		\end{figure}

\begin{figure}[h!]
			\subfloat[\label{fig:OA1_2}]{\includegraphics[width=\textwidth]{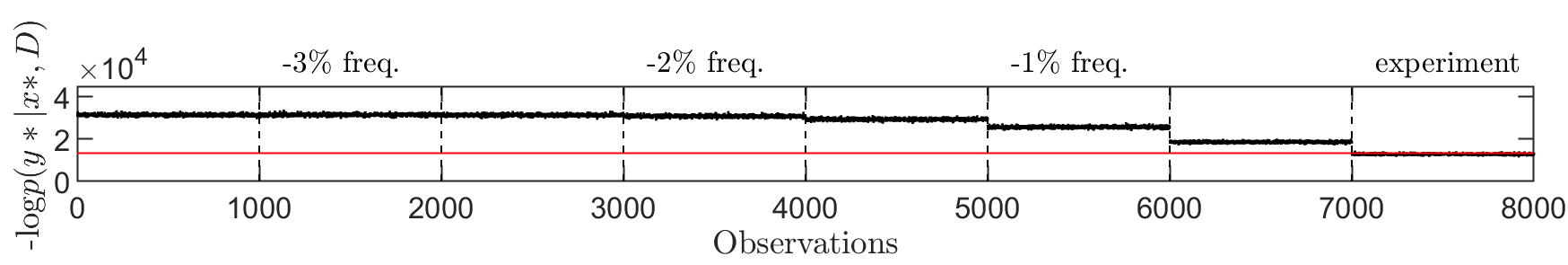}} \\[-0.25ex]
			\subfloat[\label{fig:OA2_2}]{\includegraphics[width=\textwidth,trim = {0cm 0cm 0cm 0.2cm},clip]{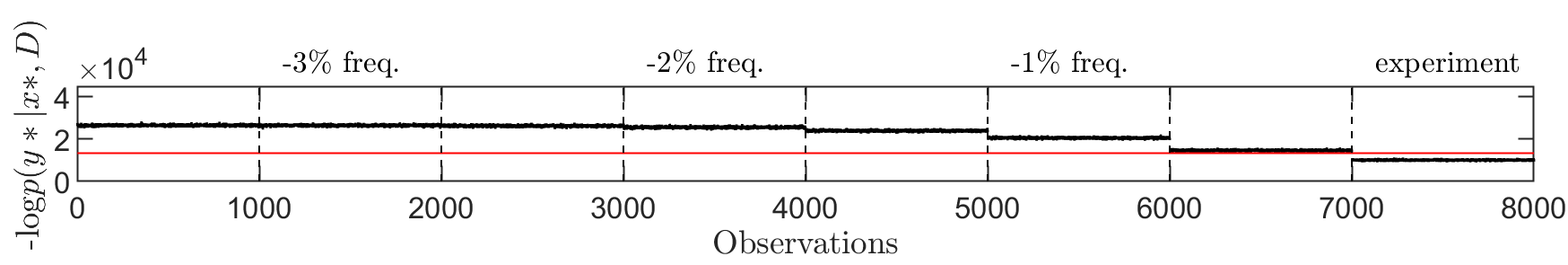}} \\[-0.25ex]
			\subfloat[\label{fig:OA3_2}]{\includegraphics[width=\textwidth,trim = {0cm 0cm 0cm 0.2cm},clip]{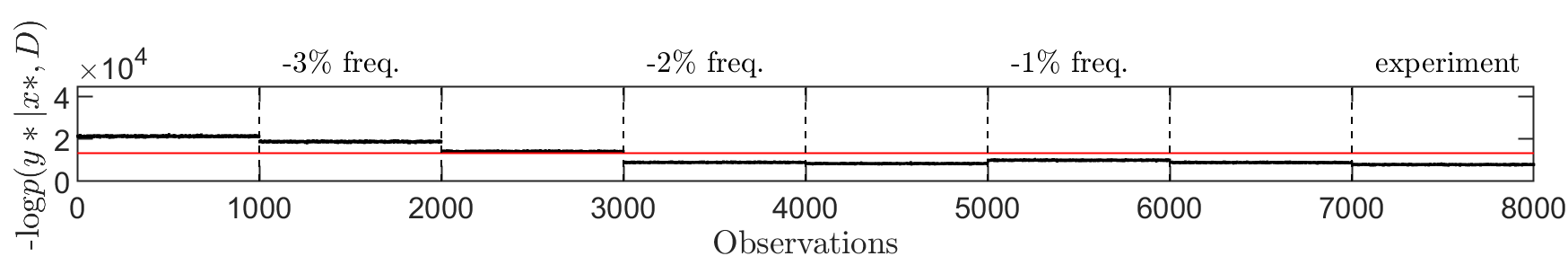}} \\[-0.25ex]
			\subfloat[\label{fig:OA4_2}]{\includegraphics[width=\textwidth,trim = {0cm 0cm 0cm 0.2cm},clip]{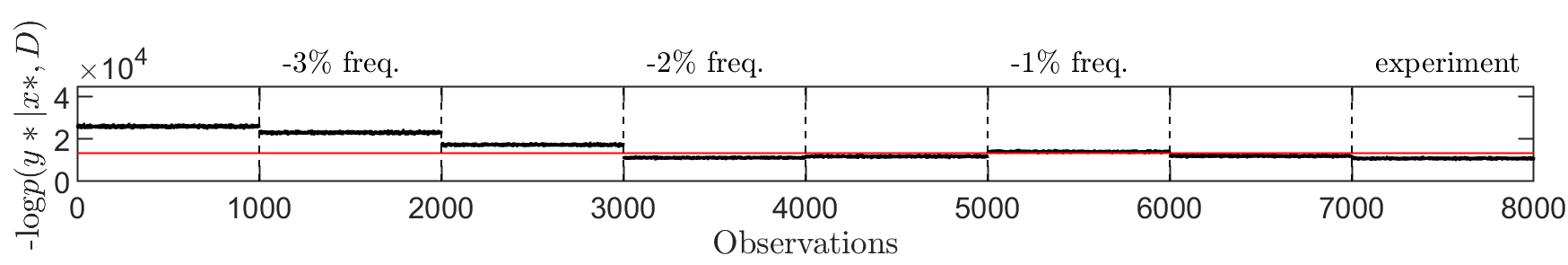}}
			\caption{Negative log marginal likelihood computed using experimental and simulated damaged results as test data for (a) Blade 1, (b) Blade 2, (c) Blade 3, and (d) \mbox{Blade 4}.}
			\vspace{-0.5cm}
		\end{figure}

	To visualise the uncertainty distribution of the FRF magnitude, the posterior predictive distributions of the real and imaginary GPs were sampled 10,000 times, and the FRF magnitude was computed from the samples. Figures \ref{fig:Synth1_2}-\ref{fig:Synth4_2} show the mean and uncertainty bounds for the OMGP magnitude, along with the magnitude of the simulated FRFs. Figures \ref{fig:OA1_2}-\ref{fig:OA4_2} show the negative log marginal likelihood computed using the measured data from the fixed-free tests as test data and the various synthesised FRFs as test data for Blades 1-4, respectively.
				
	Examining the evidence from right to left to correspond with the downward frequency shift of the FRFs, Figure \ref{fig:OA1_2} shows that -log$p(\mathbf{y}_*|\mathbf{x}_*,\mathcal{D})$ computed using replicated experimental FRFs for Blade 1 approached but did not surpass the threshold, which is expected, given that the peak belonging to Blade 1 (as shown in Figure \ref{fig:Synth1_2}) was located at the edge of the training data with respect to frequency. As the natural frequency of the synthesised FRFs decreased, the novelty index became increasingly outlying. In Figure \ref{fig:OA2_2}, similar behaviour for Blade 2 is visible, although the novelty index was further below the threshold for each case as the FRF from Blade 2 had a lower magnitude than that of Blade 1, as shown in Figures \ref{fig:Synth1_2} and \ref{fig:Synth2_2}. Figure \ref{fig:OA3_2} shows that for Blade 3, the novelty index gradually increased until approaching the threshold at -1.5\%, which corresponded to the space between the peak groupings. As the natural frequency of the simulated FRFs for Blade 3 decreased further (-1.5\% to -2\%), the novelty index fell, because the simulated FRFs aligned with the natural frequencies of Blades 1 and 2. As the natural frequency of the synthesised FRFs continued to decrease (-2.5\% to -3.5\%), the novelty index became increasingly outlying. Figure \ref{fig:OA4_2} shows similar results for Blade 4; however, because Blade 4 had a larger FRF magnitude than Blade 3, the novelty index was closer to the threshold. These results suggest that for Blades 1 and to a lesser extent Blade 3, the technique would be quite sensitive to downward frequency shifts. The technique would be less sensitive to frequency reductions for Blades 3 and 4, as a decrease in frequency may result in overlap with the normal-condition peaks for Blades 1 and 2. Note that this metric was computed in a functional sense, meaning that the test data included the full (band-limited) FRF, rather than individual test points, and the full covariance matrices of the OMGP were used to compute the novelty index. Considering the metric in a functional sense means that if the test data fit a function significantly different than the form, then the data could be flagged as outlying regardless of whether the majority of individual test points were within the variance bounds.

\section{Concluding remarks}
	
	In this work, a generalised model, called a \emph{form}, was developed for a homogeneous population of structures using a mixture of probabilistic regression models, to account for the horizontal variability among the members. In addition, a practical implementation of the technology was presented, where simulated damaged data were compared against the form to evaluate model performance in a novelty detection application.
		
\section*{Acknowledgements}
	The authors gratefully acknowledge the support of the UK Engineering and Physical Sciences Research Council (EPSRC), via grant references EP/R003645/1 and EP/R004900/1. L.A.\ Bull was supported by Wave 1 of the UKRI Strategic Priorities Fund under the EPSRC grant EP/W006022/1, particularly the \textit{Ecosystems of Digital Twins} theme within that grant and the Alan Turing Institute.
	
	This research made use of The Laboratory for Verification and Validation (LVV), which was funded by the EPSRC (grant numbers EP/J013714/1 and EP/N010884/1), the European Regional Development Fund (ERDF) and the University of Sheffield. The authors would like to extend special thanks to Robin Mills, for supporting this project via experimental design and setup, and to Michael Dutchman, for assisting with experimental setup.
	
\bibliographystyle{splncs04} 
\bibliography{references}

\end{document}